\documentclass[10pt,twocolumn,letterpaper]{article}
\usepackage{cuted}
\usepackage{capt-of} 
\usepackage{times}
\usepackage{epsfig}
\usepackage{graphicx}
\usepackage{iccv}
\usepackage{amsmath}
\usepackage{amssymb}
\usepackage{amssymb}
\usepackage{pifont}
\usepackage{booktabs} 
\usepackage{subcaption}
\newcommand{\cmark}{\ding{51}}
\newcommand{\xmark}{\ding{55}}

\usepackage[export]{adjustbox}
\usepackage[title]{appendix}
\usepackage{lipsum}

\newcommand\blfootnote[1]{
  \begingroup
  \renewcommand\thefootnote{}\footnote{#1}
  \addtocounter{footnote}{-1}
  \endgroup
}

\usepackage[pagebackref=true,breaklinks=true,letterpaper=true,colorlinks,bookmarks=false]{hyperref}

\iccvfinalcopy
\ificcvfinal\fi

\begin{document}

\title{Unconstrained Scene Generation with Locally Conditioned Radiance Fields}

\author{Terrance DeVries$^\dagger$\textsuperscript{1} \quad \qquad Miguel Angel  Bautista\textsuperscript{1} \quad \qquad Nitish Srivastava\textsuperscript{1}
\and
Graham W. Taylor\textsuperscript{2,3}
\and
Joshua M. Susskind\textsuperscript{1}
\and
\textsuperscript{1}Apple\qquad
\textsuperscript{2}University of Guelph\qquad
\textsuperscript{3}Vector Institute
}
 
\maketitle
\ificcvfinal\thispagestyle{empty}\fi
 
\begin{strip}
\centering
\vspace{-4em}
\includegraphics[width=\textwidth]{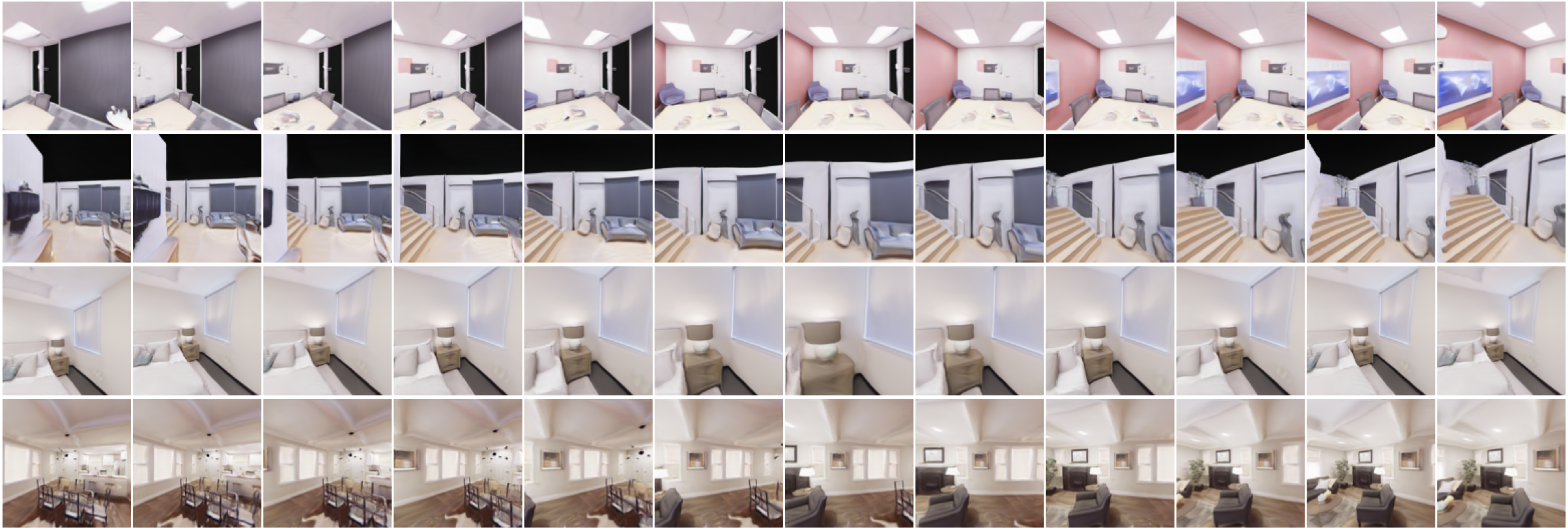}
\captionof{figure}{Scenes sampled from our learned prior, rendered from freely moving camera paths containing various rotation, translation, and forward/backward motions.}
\label{fig:flythrough}
\end{strip}

\begin{abstract}
\vspace{-1.0em}
We tackle the challenge of learning a distribution over complex, realistic, indoor scenes. In this paper,
we introduce \textbf{Generative Scene Networks} (GSN), which learns to decompose scenes into a collection of many local radiance fields that can be rendered from a free moving camera. Our model can be used as a prior to generate new scenes, or to complete a scene given only sparse 2D observations.
Recent work has shown that generative models of radiance fields can capture properties such as multi-view consistency and view-dependent lighting. However, these models are specialized for constrained viewing of single objects, such as cars or faces.
Due to the size and complexity of realistic indoor environments, existing models lack the representational capacity to adequately capture them.
Our decomposition scheme scales to larger and more complex scenes while preserving details and diversity, and the learned prior enables high-quality rendering from viewpoints that are significantly different from observed viewpoints. When compared to existing models, GSN produces quantitatively higher-quality scene renderings  across several different scene datasets. 
\end{abstract}

\begin{table*}[t]
\small
    \centering
    \begin{tabular}{cccccccc}
    \toprule
    Model & Multiple Scenes/Objects & Generative & Latent  & Radiance Field & Scene Level & Camera Placement\\
    \midrule 

    NeRF \cite{nerf} & \xmark & \xmark & - & \cmark & \cmark & Sphere/Wide-baseline\\ 
    NSVF \cite{nsvf} & \xmark & \xmark & 3D & \cmark & \cmark & Sphere/Wide-baseline \\ 
    PixelNerf \cite{pixelnerf} & \cmark & \xmark & N$\times$2D & \cmark & \cmark & Sphere/Wide-baseline \\
    GRAF \cite{graf} & \cmark & \cmark & 1D & \cmark & \xmark & Sphere  \\ 
    HoloGAN \cite{hologan} & \cmark & \cmark & 1D & \xmark & \xmark & Sphere \\ 
    PlatonicGAN \cite{platonicgan} & \cmark & \xmark & 1D & \xmark & \xmark & Sphere \\ 
    ENR \cite{enr} & \cmark & \xmark  & 3D & \xmark & \xmark & Sphere \\ 
    GTM-SM \cite{gtmsm} & \cmark & \cmark & 1D & \xmark & \cmark & Free moving\\ 
    ISS \cite{iss} & \cmark & \xmark & 2D & \xmark & \cmark & Free moving\\ 
    GSN (ours)  & \cmark & \cmark & 2D & \cmark & \cmark & Free moving\\ 
    \bottomrule
    \end{tabular}
    \caption{Summary of contributions and comparison with relevant related work. (\textbf{Multiple Scene/Objects}): Ability to model multiple scenes/objects in the same network. (\textbf{Generative}) Whether the model is generative (\eg allows for free sampling).  (\textbf{Latent}) Latent code spatial dim. (\textbf{Radiance Field}) Whether the model predicts a radiance field. (\textbf{Scene level}) Results demonstrated in scene-level environments. (\textbf{Camera Placement}) What camera motion is permitted?}
    \label{tab:contributions}
\end{table*}
\section{Introduction}
\vspace{-1em}

\blfootnote{$\dagger$ Work done as part of an internship at Apple.}

Spatial understanding entails the ability to infer the geometry and appearance of a scene when observed from any viewpoint or orientation given sparse or incomplete observations. Although a wide range of geometry and learning-based approaches for 3D view synthesis \cite{extremeviewsynthesis,mpi1,llff,nerf,derf,mpi2,mpi3} can interpolate between observed views of a scene, they cannot extrapolate to infer unobserved parts of the scene. The fundamental limitation of these models is their inability to learn a prior over scenes. As a result, learning-based approaches have limited performance in the extrapolation regime, whether it be inpainting disocclusions or synthesizing views beyond the boundaries of the given observations. For example, the popular NeRF \cite{nerf} approach represents a scene via its radiance field, enabling continuous view interpolation given a densely captured scene. However, since NeRF does not learn a scene prior, it cannot extrapolate views. On the other hand, conditional auto-encoder models for view synthesis \cite{gqn,appearanceflow,multiview2novelview,enr,tbn,tatarchenko} are able to extrapolate views of simple objects from multiple viewpoints and orientations. Yet, they overfit to viewpoints seen during training (see \cite{monocularviewsyn} for a detailed analysis). Moreover, conditional auto-encoders tend to favor point estimates (\eg the distribution mean) and produce blurry renderings when extrapolating far from observed views as a result. %

A learned prior for scenes may be used for \textit{unconditional} or \textit{conditional} inference. A compelling use case for unconditional inference is to generate realistic scenes and freely move through them in the absence of input observations, relying on the prior distribution over scenes (see Fig.~\ref{fig:flythrough} for examples of trajectories of a freely moving camera on scenes sampled from our model). Likewise, conditional inference lends itself to different types of problems. For instance, plausible scene completions may be sampled by inverting scene observations back to the learned scene prior \cite{ganinversion}. A generative model for scenes would be a practical tool for tackling a wide range of problems in machine learning and computer vision, including model-based reinforcement learning \cite{worldmodels}, SLAM \cite{neuralslam2,neuralslam1}, content creation~\cite{stylegan}, and adaptation for AR/VR or immersive 3D photography.

In this paper we introduce \textit{Generative Scene Networks} (GSN), a generative model of scenes that allows view synthesis of a freely moving camera in an open environment. Our contributions are the following. We: \textit{(i)} introduce the first generative model for unconstrained scene-level radiance fields; \textit{(ii)} demonstrate that decomposing a latent code into a grid of locally conditioned radiance fields results in an expressive and robust scene representation, which outperforming strong baselines; \textit{(iii)} infer observations from arbitrary cameras given a sparse set of observations by inverting GSN (\ie, fill in the scene); and \textit{(iv)} show that GSN can be trained on multiple scenes to learn a rich scene prior, while rendered trajectories are smooth and consistent, maintaining scene coherence.

\begin{figure*}[t]
    \centering
    \includegraphics[width=0.9\textwidth]{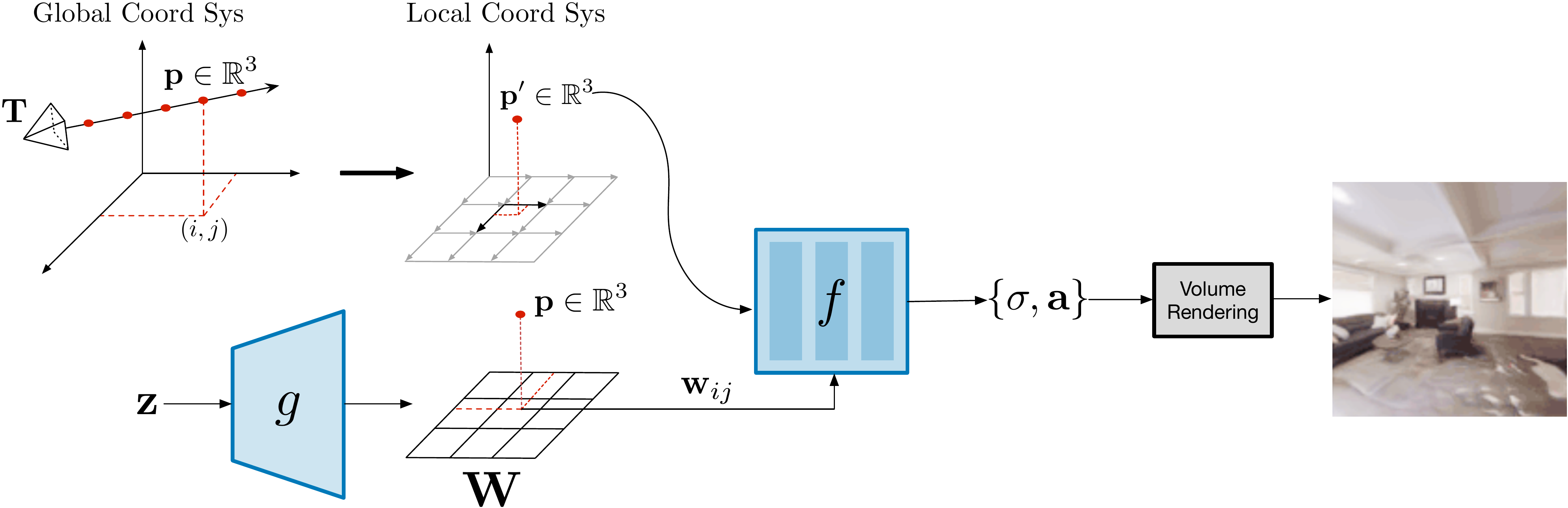}
    \caption{Architecture of the GSN generator. We sample a latent code $\textbf{z}~\sim~p_z$ that is fed to our global generator $g$ producing a local latent grid $\textbf{W}$. This local latent grid $\textbf{W}$ conceptually represents a latent scene ``floorplan'' and is used for locally conditioning a radiance field $f$ from which images are rendered via volumetric rendering. For a given point $\textbf{p}$ expressed in a global coordinate system to be rendered, we sample $\textbf{W}$ at the location $(i, j)$, given by $\textbf{p}$ resulting in $\textbf{w}_{ij}$. In turn $f$ takes as input $\textbf{p}'$ which results from expressing $\textbf{p}$ relative to the local coordinate system of $\textbf{w}_{ij}$.}
    \label{fig:gsn_model}
\end{figure*}

\section{Related Work}

An extensive body of literature has tackled view synthesis of simple synthetic objects \cite{shapenet} against a homogeneous background \cite{enr,tbn,gaf,multiview2novelview,appearanceflow}. Relevant models typically rely on predicting 2D/3D flow fields to transform pixels in the input view to pixels in the predicted view. Moreover, in \cite{srn} a continuous representation is introduced in the form of the weights of a parametric vector field, which is used to infer appearance of 3D points expressed in a world coordinate system. At inference time the model is optimized to find latent codes that are predictive of the weights. The above approaches rely on the existence of training data of the form \textit{input-transformation-output} tuples, where at training time the model has access to two views (\textit{input} and \textit{output}) in a viewing sphere and the corresponding 3D transformation. Sidestepping this requirement \cite{platonicgan,hologan} showed that similar models can be trained in an adversarial setting without access to \textit{input-transformation-output} tuples.

Recent approaches \cite{nsvf,nerf,derf,fvs,svs} have been shown to encode a continuous radiance field in the parameters of a neural network by fitting it to a collection of high resolution views of realistic scenes. The encoded radiance field is rendered using the classical volume rendering equation \cite{volumerendering} and a reconstruction loss is used to optimize the model parameters. While successful at modelling the radiance field at high resolutions, these approaches require optimizing a new model for every scene (where optimization usually takes days on commodity hardware). Thus, the fundamental limitation of these works is that they are unable to represent multiple scenes within the same model. As a result, these models cannot learn a prior distribution over multiple scenes.

Some newer works have explore generative radiance field models in the adversarial setting \cite{pigan,giraffe,graf}.
However, \cite{pigan,graf} are restricted to modeling single objects where a camera is positioned on a viewing sphere oriented towards the object, and they use a 1D latent representation, which prevents them from efficiently scaling to full scenes (c.f.~\S~\ref{sect:experiments} for experimental evidence). Concurrent to our work, Niemeyer et al.~\cite{giraffe} model a scene with multiple entities but report results on single-object datasets and simple synthetic scenes with 2-5 geometrical shapes in the CLEVR dataset \cite{clevr}. The object-level problem may be solved by learning a pixel-aligned representation for conditioning the radiance field \cite{grf,pixelnerf,sharf,ibrnet}. However, \cite{grf,pixelnerf,sharf,ibrnet} are designed as conditional auto-encoders as opposed to a generative model, thereby ignoring the fundamentally stochastic nature of the view synthesis problem. In \cite{infinitenature} a model is trained on long trajectories of nature scenarios. However, due to their iterative refinement approach the model has no persistent scene representation. While allowing for perpetual generation, the lack of a persistent scene representation limits its applicability for other downstream tasks that may require loop closure. 

Approaches tackling view synthesis for a freely moving camera in a scene offer the capability to fully explore a scene \cite{gqn,gtmsm,iss}. Irrespective of the output image resolution, this is strictly a more complex problem than a camera moving on a sphere oriented towards a single object \cite{srn,appearanceflow,multiview2novelview,pixelnerf,graf,giraffe,pigan}. For scenes with freely moving cameras the model needs to learn to represent a full scene that consists of a multitude of objects rather than a single object. In this setup, it is often useful to equip models with a memory mechanism to aggregate the set of incoming observations. In particular, it has been useful to employ a dictionary-based memory to represent an environment where keys are camera poses and values are latent observation representations \cite{gtmsm}. Furthermore, the memory may be extended to a 2D feature map that represents at top-down view of the environment \cite{gqn,iss}. At inference time, given a query viewpoint the model queries the memory and predicts the expected observation. This dependency on stored observations significantly limits these models' ability to cope with unseen viewpoints. GSN can perform a similar task by matching observations to scenes in its latent space, but with the added benefit that the learned scene prior allows for extrapolation beyond the given observations. A summarized comparison of GSN with the most relevant related work is shown in Tab. \ref{tab:contributions}.

\section{Method}

 As in traditional GANs \cite{gan}, our model is composed of a generator $G_\theta$ and a discriminator $D_\phi$, with latent codes $\textbf{z}~\sim~p_z$, where $p_z$ denotes the prior distribution. The generator $G_\theta$ is composed of two subnetworks $G~=~g_{\theta_g}~\cdot~f_{\theta_f}$: a global generator $g_{\theta_g}$ and a locally conditioned radiance field network $f_{\theta_f}$ (in the rest of the text and figures we drop the subscripts for clarity). Unlike standard 2D GANs, which often only require samples $\textbf{z}\sim~p_z$ as input, GSN also takes a camera pose $\textbf{T}~\in~SE(3)$ from an empirical pose distribution $p_T$, as well as camera intrinsic parameters $\textbf{K}~\in~\mathbb{R}^{3~\times~3}$. These additional inputs allow for explicit control over the viewpoint and field of view of the output image, which is rendered via a radiance field. Fig.~\ref{fig:gsn_model} outlines the GSN network architecture.

\subsection{Global Generator}

In GRAF \cite{graf} and $\pi$-GAN \cite{pigan} a 1D global latent code $\textbf{z}\sim p_z$ is used to condition an MLP which parameterizes a single radiance field. Although such 1D global latent code can be effective when representing individual object categories, such as cars or faces, it does not scale well to large, densely populated scenes. Simply increasing the capacity of a single radiance field network has diminishing returns with respect to render quality \cite{derf}. It is more effective to distribute a scene among several smaller networks, such that each can specialize on representing a local region \cite{derf}. Inspired by this insight we propose the addition of a global generator that learns to decompose large scenes into many smaller, more manageable pieces.

\begin{figure}[t]
    \centering
    \subcaptionbox{\centering Global generator}{\includegraphics[width=0.22\textwidth]{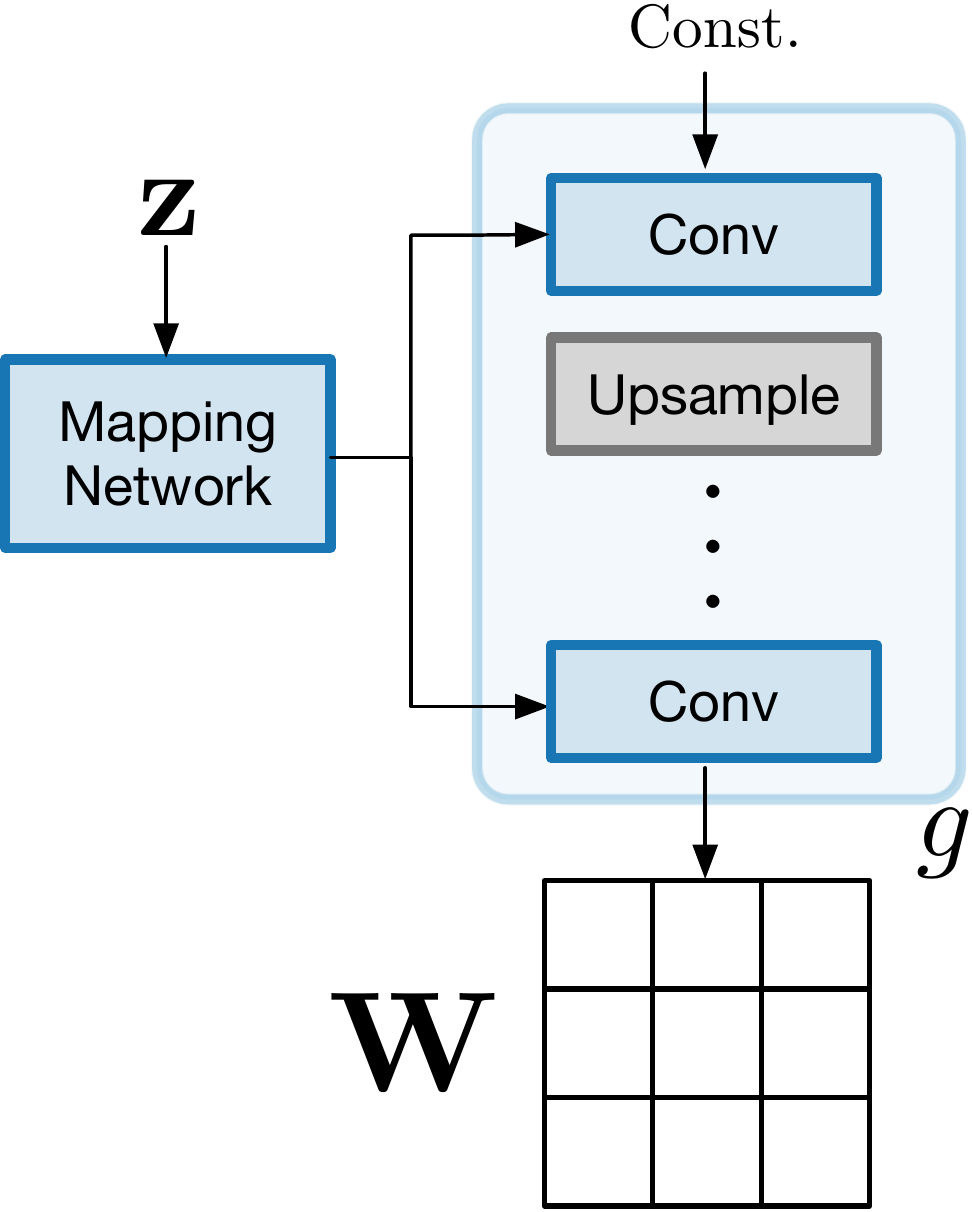}}
    \hspace{0.4cm}
    \subcaptionbox{\centering Locally conditioned radiance field network}{
    \includegraphics[width=0.17\textwidth]{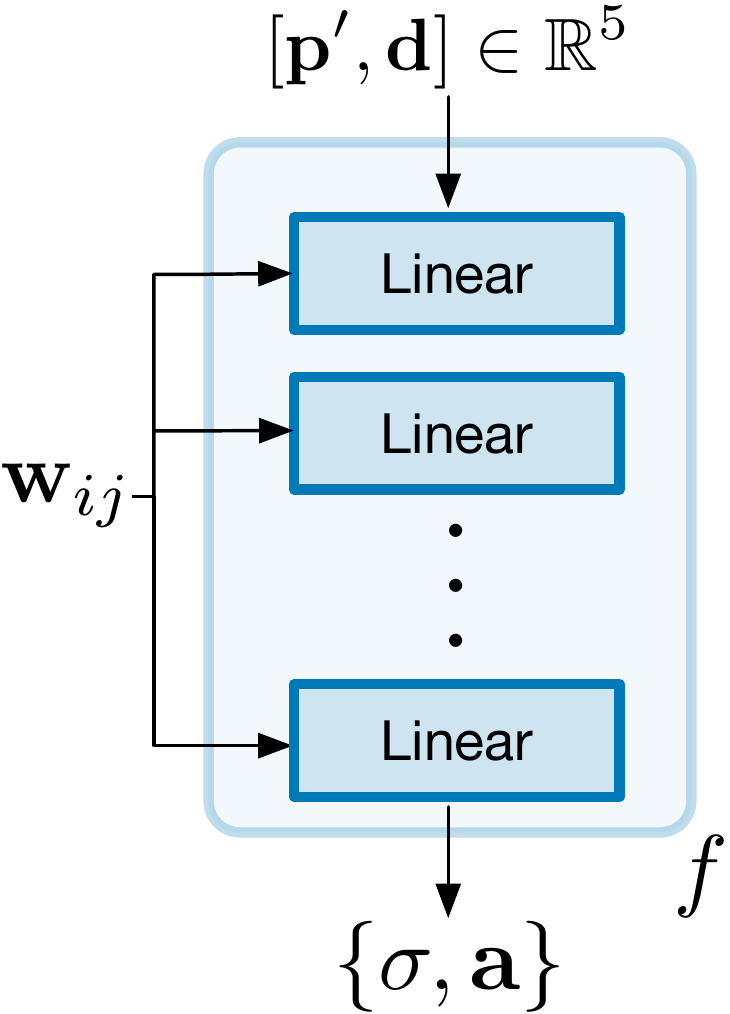}}\\
    \vspace{0.3cm}
    \caption{(a) Architecture of global generator $g$. We use a mapping network, modulated convolutional blocks, and a learned constant input as in StyleGAN2 \cite{stylegan2}. (b) Architecture of the locally conditioned radiance field network $f$. Latent code $\textbf{w}_{ij}$, sampled from $\textbf{W}$, is used to modulate linear layers, similar to CIPS~\cite{modfc}.}
    \label{fig:global_generator}
\end{figure}

The global generator $g$ maps from the global latent code $\textbf{z} \in  \mathbb{R}^d $, which represents an entire scene, to a 2D grid of local latent codes $\textbf{W} \in \mathbb{R}^{c \times s \times s}$, each of which represent a small portion of the larger scene (see Fig.~\ref{fig:global_generator}a). Conceptually, the 2D grid of local latent codes can be interpreted as a latent \textit{floorplan} representation of a scene, where each code is used to locally condition a radiance field network. We opt for a 2D representation produced with a convolutional generator for computational efficiency, but our framework can be easily extended to a 3D grid~\cite{nsvf}. 

\subsection{Locally Conditioned Radiance Field}\label{sec:local_radiance_fields}

To render an image $\hat{\textbf{X}} \in \mathbb{R}^{3 \times w \times h}$ given a grid of latent codes $\textbf{W} \in \mathbb{R}^{c \times s \times s}$, and a camera with pose $\textbf{T}=[\textbf{R}, \textbf{t}] \in SE(3)$ and intrinsics $\textbf{K} \in \mathbb{R}^{3 \times 3}$,  we model scenes with a locally conditioned radiance field \cite{nerf} $f:\mathbb{R}^5 \times \mathbb{R}^c \rightarrow [0, 1] \times \mathbb{R}^3$. To predict radiance for a 3D point $\textbf{p}$ and 2D camera direction $\textbf{d}$, $f$ is locally conditioned on a vector $\textbf{w}_{ij} \in \mathbb{R}^c$ sampled at a discrete location $(i, j)$ from the grid of latent codes $\textbf{W}$ using bi-linear sampling \cite{stn}. The location $(i, j)$ is naturally given by the projection of $\textbf{p}$ on the $zx$-plane (assuming the $y$-axis points up).

Radiance fields are usually defined over $\mathbb{R}^3$ considering a global coordinate system \cite{nerf,graf,pigan} (\ie a coordinate system that spans the whole scene/object). This has been successful since these works do not learn a decomposition of scenes into multiple radiance fields. To decompose the scene into a grid of \textit{independent} radiance fields, we enable our model to perform spatial sharing of local latent codes; the same latent code $w_{ij}$ can be used to represent the same scene part irrespective of its position $(i, j)$ on the grid (see Fig.~\ref{fig:coordinate_system} for empirical results). In order to achieve this spatial sharing, it is not sufficient to learn a grid of local latent codes, it is also necessary to decompose the global coordinate system into multiple local coordinate systems (one for each local latent $\textbf{w}_{ij}$). The reason for this is to prevent $f$ from relying on the absolute coordinate value of input points to predict radiance. As a result, the input to $f$ is  $\textbf{p}'$, which results from expressing $\textbf{p}$ relative to the local coordinate system of its corresponding local latent $\textbf{w}_{ij}$ and applying positional encoding~\cite{nerf}.

Finally, the locally conditioned radiance field $f$ outputs two variables: the occupancy $\sigma \in [0, \infty]$ and the appearance $\textbf{a} \in \mathbb{R}^3$ (see Fig.~\ref{fig:global_generator}b). To render $\hat{\textbf{X}}$ given $f$ we leverage Eq.~\ref{eq:rendering} for implicit volumetric rendering \cite{nerf} and densely evaluate $f$ on points uniformly sampled along ray directions $\textbf{r}$, where each pixel in $\hat{\textbf{X}}$ corresponds to a ray and the color $\textbf{c} \in \mathbb{R}^3$ of pixel/ray $\textbf{r}$ is obtained by approximating the integral in Eq. \ref{eq:rendering}. For computational efficiency, we use Eq.~\ref{eq:rendering}  to render a feature map as in \cite{giraffe}. The rendered feature map is then is upsampled with a small convolutional refinement network to achieve the desired output resolution, resulting in final rendered output $\hat{\textbf{X}}$.

\begin{eqnarray}
\mathbf{c}(\mathbf{r}, \textbf{W}) = \int_{u_n}^{u_f} Tr(u) \sigma\left(\mathbf{r}(u), \textbf{w}_{ij}\right) \mathbf{a}\left(\mathbf{r}(u), \mathbf{d}, \textbf{w}_{ij}\right) du \nonumber  \\
Tr(u) = \exp \left(-\int_{u_n}^{u} \sigma(\textbf{r}(u), \textbf{w}_{ij}) du\right)
\label{eq:rendering}
\end{eqnarray}

\subsection{Sampling Camera Poses}\label{sec:camera_sampling}

Unlike standard 2D generative models which have no explicit concept of viewpoint, radiance fields require a camera pose for rendering an image. Therefore, camera poses $\textbf{T}=[\textbf{R} | \textbf{t}] \in SE(3)$ need to be sampled from pose distribution $p_T$ in addition to the latent code $\textbf{z}~\sim~p_z$, which is challenging for the case of realistic scenes and a freely moving camera. GRAF~\cite{graf} and $\pi$-GAN~\cite{pigan} avoid this issue by training on datasets containing objects placed at the origin, where the camera is constrained to move on a viewing sphere around the object and oriented towards the origin.\footnote{Camera poses $\textbf{T}$ on a sphere oriented towards the origin are constrained to $SO(3)$ as opposed to free cameras in $SE(3)$.} For GSN, sampling camera poses becomes more challenging due to i) the absence of such constraints (\ie need to sample $\textbf{T} \in SE(3)$), ii) and the possibility of sampling invalid locations, such as inside walls or other solid objects that sporadically populate the scene.

To overcome the issue of sampling invalid locations we perform stochastic weighted sampling over a an empirical pose distribution $p_T$ composed by a set of candidate poses, where each pose is weighted by the occupancy (i.e., the $\sigma$ value predicted by the model) at that location. When sampling the candidate poses, we query the generator at each candidate location to retrieve the corresponding occupancy. Then the occupancy values are normalized with a softmin to produce sampling weights for a multinomial distribution. %
This sampling strategy reduces the likelihood of sampling invalid camera locations while retaining stochasticity required for scene exploration and sample diversity.

\subsection{Discriminator}

Our model adopts the convolutional discriminator architecture from StyleGAN2~\cite{stylegan2}.%
The discriminator takes as input an image, concatenated with corresponding depth map normalized between $[0, 1]$ 
and predicts whether the input comes from the true distribution or the generated distribution. We found it critical to add a small decoder $C$ to the discriminator and to enforce a reconstruction penalty on real images, similar to the self-supervised discriminator proposed by Lui et al.~\cite{self_supervised_discriminator}. The additional regularization term restricts the discriminator from overfitting, while encouraging it to learn relevant features about the input that will provide a useful training signal for the generator.

\subsection{Training}

Let $\mathcal{X} = \{\textbf{X}\}: \textbf{X} \in \mathbb{R}^{4 \times w \times h}$ denote the set of  i.i.d RGB-D samples obtained by recording camera trajectories on the true distribution of scenes $p_S$. Generator $G$, discriminator $D$, and decoder $C$ are parameterized by $\theta$, $\phi$, and $\omega$, respectively.  We train our model by optimizing the non-saturating GAN loss~\cite{gan} with R1 gradient penalty~\cite{r1}, as well as the discriminator reconstruction objective~\cite{self_supervised_discriminator}: 
\newcommand{\generator}{G_{\theta}}
\newcommand{\discriminator}{D_{\phi}}
\newcommand{\decoder}{C_{\omega}}
\begin{equation}
\begin{split}
\mathcal{L}(\theta, \phi, \omega) & = \mathbf{E}_{\mathbf{z}\sim p_z, \textbf{T}\sim p_T}[h(\discriminator(\generator(\mathbf{z}, \textbf{T})))]
\\
& + \mathbf{E}_{\mathbf{X}\sim p_\mathcal{S}}[h(-\discriminator(\mathbf{X})) + \lambda_{R1} \lvert \nabla \discriminator (\mathbf{X}) \rvert ^ 2 
\\
& + \lambda_{Recon} \lvert \mathbf{X} - \decoder(\discriminator(\mathbf{X}))  \rvert ^ 2] ,
\\
& \text{where} \quad h(u) = -\log(1 + \exp(-u)).
\end{split}
\end{equation}

\section{Experiments} \label{sect:experiments}
In this section we report both quantitative and qualitative experimental results on different settings. First, we compare the generative performance of GSN with recent state-of-the-art approaches. Second, we provide extensive ablation experiments that show the quantitative improvement obtained by the individual components of our model. Finally, we report results on view synthesis by inverting GSN and querying the model to predict views of a scene given a set of input observations. 

\begin{figure*}[t]
    \centering
        \begin{subfigure}{0.35\linewidth}
            \centering
            \includegraphics[width=\linewidth]{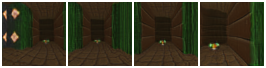}
            \includegraphics[width=\linewidth]{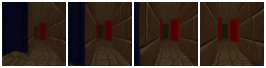}
            \includegraphics[width=\linewidth]{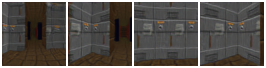}
        \end{subfigure} 
        \hspace{1cm}
        \begin{subfigure}{0.35\linewidth}
            \centering
            \includegraphics[width=\linewidth]{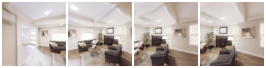}
            \includegraphics[width=\linewidth]{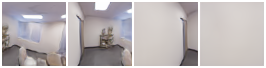}
            \includegraphics[width=\linewidth]{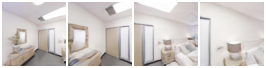}
        \end{subfigure} \hfill
    \caption{Random trajectories through scenes generated by GSN. Models are trained on VizDoom \cite{vizdoom} (left), Replica \cite{replica} (right) at $64\times64$ resolution. We omit qualitative results for AVD \cite{avd} due to unclear licensing terms regarding reproduction of figures for this dataset.}
    \label{fig:scene_samples}
\end{figure*}

\begin{table*}[!t]
\small
\begin{center}
 \begin{tabular}{lcccccc}
    \toprule
    &  \multicolumn{2}{c}{VizDoom \cite{vizdoom}} & \multicolumn{2}{c} {Replica \cite{replica}} & \multicolumn{2}{c} {AVD \cite{avd}} \\
   \cmidrule(r){2-3}\cmidrule(r){4-5}\cmidrule(r){6-7}
  & FID $\downarrow$ & SwAV-FID $\downarrow$ & FID $\downarrow$ & SwAV-FID $\downarrow$ & FID $\downarrow$& SwAV-FID $\downarrow$  \\ 
 \midrule
  GRAF \cite{graf} &  $47.50 \pm 2.13$ & $5.44 \pm 0.43$ & $65.37 \pm 1.64$ & $5.76 \pm 0.14$  & $62.59 \pm 2.41$ & $6.95 \pm 0.15$ \\
  $\pi$-GAN\cite{pigan} &  $143.55 \pm 4.81$  & $15.26 \pm 0.15$ & $166.55 \pm 3.61$ & $13.17 \pm 0.20$ & $98.76 \pm 1.49$ & $9.54 \pm 0.29$ \\
  GSN (Ours) & $\mathbf{37.21 \pm 1.17}$ & $\mathbf{4.56 \pm 0.19}$  & $\mathbf{41.75 \pm 1.33}$ & $\mathbf{4.14 \pm 0.02}$ & $\mathbf{51.11 \pm 1.37}$ & $\mathbf{6.59 \pm 0.03}$ \\
  \bottomrule
\end{tabular} 
\end{center}
\caption{Generative performance of state-of-the-art approaches for generative modelling of radiance fields on 3 scene-level datasets: Vizdoom \cite{vizdoom}, Replica \cite{replica} and Active Vision (AVD) \cite{avd}, according to FID~\cite{fid} and SwAV-FID \cite{swavfid} metrics.}
\label{tab:generative_quant}
\end{table*}

\subsection{Generation Performance}

We evaluate the generative performance of our model on three datasets: i) the VizDoom environment \cite{vizdoom}, a \textit{synthetic} computer-generated world, ii) the Replica dataset \cite{replica} containing 18 scans of \textit{realistic} scenes that we render using the Habitat environment \cite{habitat}, and iii) the Active Vision Dataset (AVD)~\cite{avd} consisting of 20k images with noisy depth measurements from 9 \textit{real world} scenes.\footnote{We will release 
code for reproducibility.} Images are resized to $64\times64$ resolution for all generation experiments. To generate data for the VizDoom and Replica experiments we render $100$ sequences of $100$ steps each, where an interactive agent explores a scene collecting RGB and depth observations as well as camera poses. Sequences for AVD are defined by an agent performing a random walk through the scenes according to rules defined in~\cite{iss}.
At training time we sample sub-sequences and express camera poses relative  the middle frame of the sub-sequence. This normalization enforces an egocentric coordinate system whose origin is placed at the center of $\textbf{W}$ (see supplementary material for details).

We compare GSN to two recent approaches for generative modelling of radiance fields: GRAF \cite{graf} and $\pi$-GAN \cite{pigan}. For fair comparison all models use the same base architecture and training pipeline, with two main differences between models. The first difference is that GSN uses locally conditioned radiance fields, while GRAF and $\pi$-GAN use global conditioning. The second difference is the type of layer used in the radiance field generator network: GRAF utilizes linear layers, $\pi$-GAN employs modulated sine layers, and GSN uses modulated linear layers. Quantitative performance is evaluated with the Fr\'{e}chet Inception Distance (FID)~\cite{fid} and SwAV-FID \cite{swavfid} metrics, which measure the distance between the distributions of real and generated images in pretrained image embedding spaces. We sample 5,000 real and 5,000 generated images when calculating either of these metrics.

Tab. \ref{tab:generative_quant} shows the results of our generative modelling comparison. Despite our best attempts to tune $\pi$-GAN's hyperparameters, we find that it struggles to learn detailed depth in this setting, which significantly impedes render quality and leads to early collapse of the model. GSN achieves much better performance than GRAF or $\pi$-GAN across all datasets, obtaining an improvement of 10-14 absolute points on FID. We attribute GSN's higher performance to the increased expressiveness afforded by the locally conditioned radiance field, and not the specific layer type used (compare GRAF in Tab. \ref{tab:generative_quant} to GSN + global latents in Tab. \ref{tab:latent_code_ablation}). As a measure of qualitative results we show scenes sampled from GSN trained on the Replica dataset at $128\times128$ resolution in Fig. \ref{fig:flythrough}, and on the VizDoom, Replica, and AVD datasets at $64\times64$ resolution in Fig. \ref{fig:scene_samples}.

\begin{figure*}[t]
    \centering
    \includegraphics[width=\textwidth]{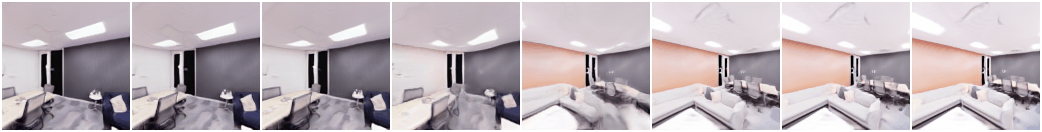}
    \includegraphics[width=\textwidth]{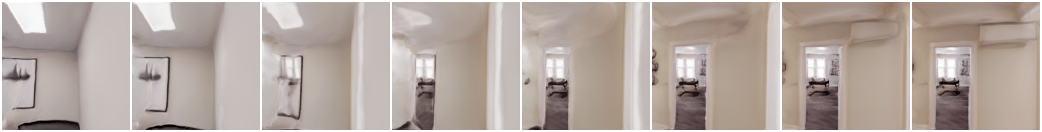}
    \caption{Two example latent interpolations between global latent codes $\textbf{z}$. Scenes transition smoothly by aligning geometry features such as walls (top) and appearance features such as the picture frame and doorway (bottom). Views are rendered from a fixed camera pose.}
    \label{fig:latent_interpolation_1}
\end{figure*}

\textbf{Latent Space Interpolation} To confirm that GSN is learning a useful scene prior we demonstrate in Fig.~\ref{fig:latent_interpolation_1} some examples of interpolation in the global latent space. The model aligns both geometry and appearance features such that traversing the latent space transitions smoothly between scenes without producing unrealistic off-manifold samples. 

\subsection{Ablation}

To further analyze our contributions, we report extensive ablation experiments where we show how our design choices affect the generative performance of GSN. For ablation, we perform our analysis on the Replica dataset \cite{replica} at 64 \texttimes 64 resolution and analyze the following factors: \textit{(i)} the choice of latent code representation (global vs local); \textit{(ii)} the effects of global and local coordinate systems on the learned representation; \textit{(iii)} the effect of sampled trajectory length; \textit{(iv)} the depth resolution needed in order to successfully learn a scene prior; \textit{(v)} the regularization applied to the discriminator.

\textbf{How useful are locally conditioned radiance fields?} We compare the effect of replacing our local latent codes $\textbf{W} \in \mathbb{R}^{c \times s \times s}$ with a global latent code $\textbf{w} \in \mathbb{R}^c$ that globally conditions the radiance field $f_{\theta_f}$. %
Generative performance can be improved by decomposing the scene into many independent locally conditioned radiance fields (Tab.~\ref{tab:latent_code_ablation}).

\begin{table}[!h]
\small
\begin{center}
 \begin{tabular}{lcc}
 \toprule
 Model & FID $\downarrow$ & SwAV-FID $\downarrow$ \\
 \midrule
 GSN + global latents & $68.42 \pm 3.88$ & $6.14 \pm 0.24$ \\
 GSN + local latents & $41.75 \pm 1.33$ & $4.14 \pm 0.02$ \\
 \bottomrule

\end{tabular}
\end{center}
\caption{Decomposition of the global latent code into a local latent grid has a drastic impact in performance.}
\label{tab:latent_code_ablation}
\end{table}

\textbf{What are the benefits of a local coordinate system?}
Enabling feature sharing in local coordinate systems to  (\eg a latent $w_{ij}$ can be used to represent the same scene part irrespective of its position $(i, j)$ on the grid). To empirically validate this hypothesis we train GSN with both \textit{global} and \textit{local} coordinate systems. We then sample from the prior obtaining 5k latent grids $\textbf{W}$ for each model. Next, we perform a simple rigid re-arrangement of latent codes in $\textbf{W}$ by applying a 2D rotation at different angles and measuring FID by using the resulting grids to predict a radiance field. A local coordinate system is significantly more robust to re-arranging local latent codes than a global one (Fig.~\ref{fig:coordinate_system}; see supplementary material for qualitative results).

\begin{figure}[t]
    \centering
    \includegraphics[width=0.48\textwidth]{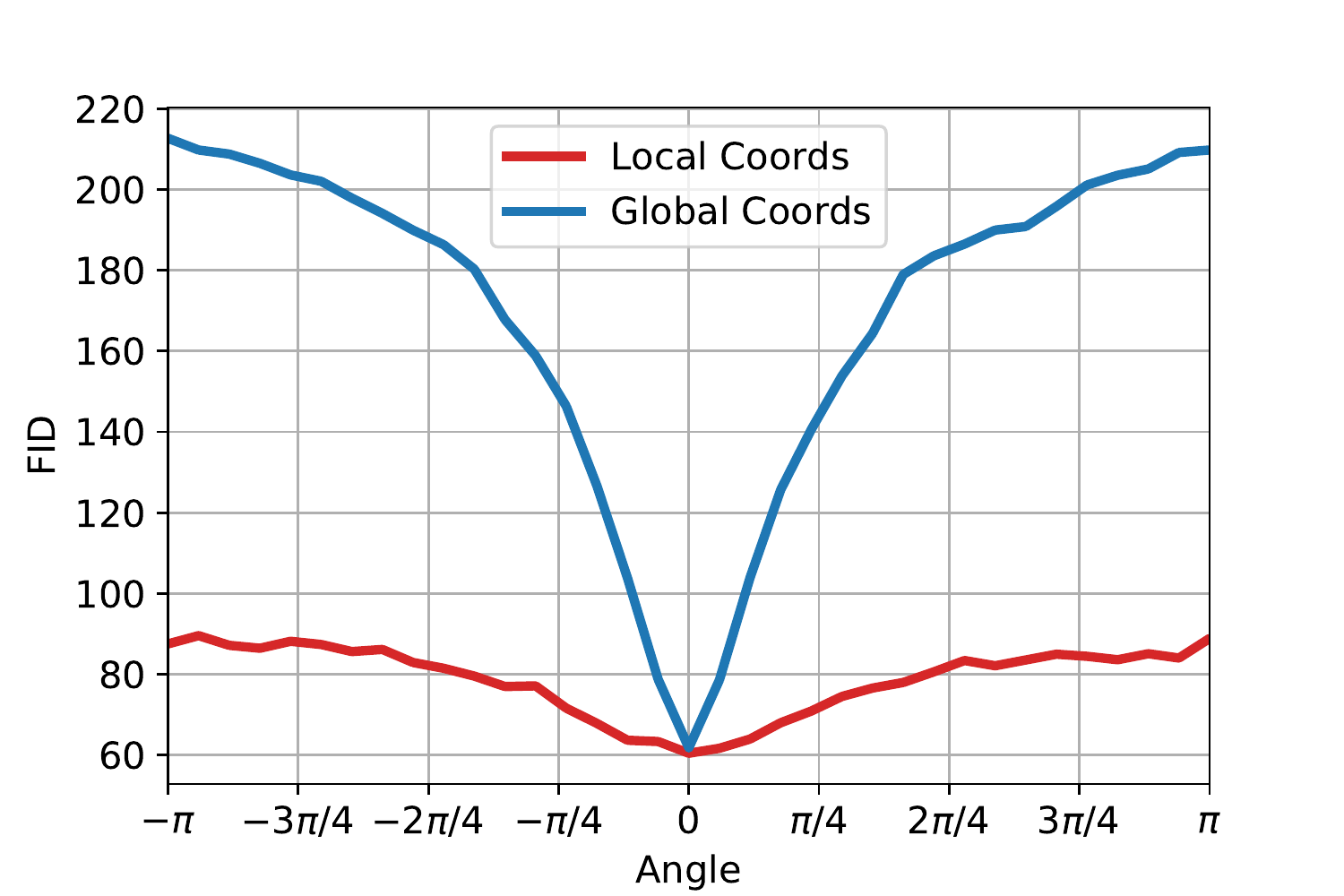}
    \caption{Robustness of generation quality \wrt to rotations of the local latent grid $\textbf{W}$. Using a local coordinate system for each local radiance field 
    limits degradation due to rigid re-arrangement of the local latent codes.
    }
    \label{fig:coordinate_system}
\end{figure}

\begin{figure}[h]
    \centering
    \includegraphics[width=0.48\textwidth]{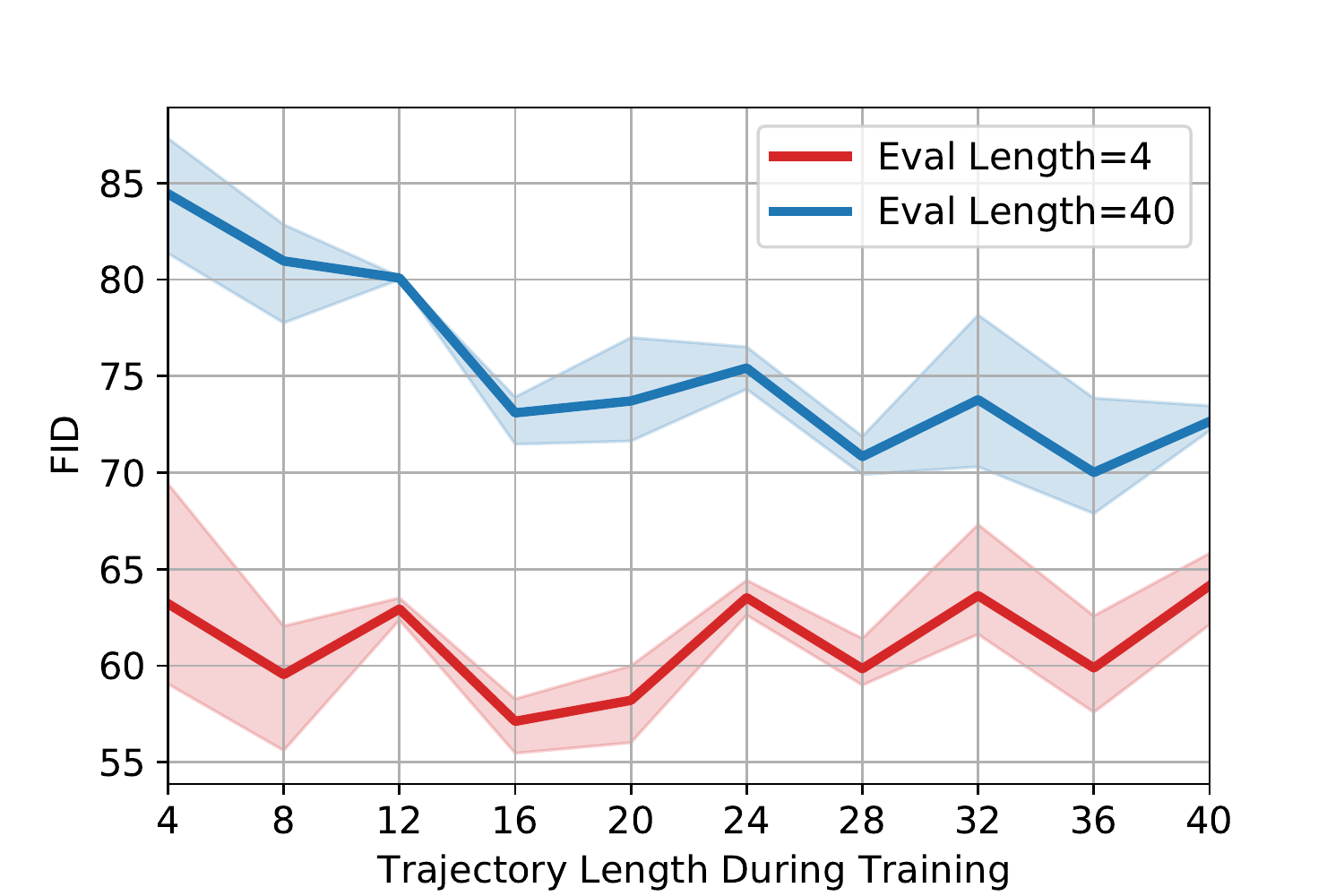}
    \caption{Effect of trajectory length (x axis) on FID scores (y axis) when evaluating short 4-step trajectories (red) vs. 40-step trajectories (blue) during training. Models trained only with short trajectories achieve good local render quality, but cannot move far from the origin without encountering quality degradation.}
    \label{fig:trajectory_length}
\end{figure}

\textbf{How long do camera trajectories need to be?} The length of trajectories that collect the camera poses affects the representation of large scenes. Given that we normalize camera poses \wrt the middle step in a trajectory, if the trajectories are too short the model will only explore small local neighbourhoods and will not be able to deal with long displacements in camera poses during test time. Models trained on short trajectories fail to generalize to long trajectories during evaluation (Fig.~\ref{fig:trajectory_length}). However, models trained with long trajectories do not struggle when evaluated on short trajectories, as evidenced by the stability of the FID when evaluated on short trajectories.

\textbf{How much depth information is required during training?} The amount of depth information used during training affects GSN. To test the sensitivity of GSN to depth information, we down-sample both real and generated depth maps to the target resolution, then up-sample back to the original resolution before concatenating with the corresponding image and feeding them to the discriminator. Without depth information GSN fails to train (Tab.~\ref{tab:depth_resolution}). However, we demonstrate that the depth resolution can be reduced to \textit{a single pixel} without finding statistically significant reduction in the generated image quality. This is a significant  result, as it enables GSN to be trained in settings with sparse depth information in future work. We speculate that in early stages of training the added depth information guides the model to learn some aspect of depth, after which it can learn without additional supervision. %

\begin{table}[h]
\small
\begin{center}
 \begin{tabular}{lcc}
 \toprule
 Depth Resolution & FID $\downarrow$ & SwAV-FID $\downarrow$ \\
 \midrule
 No depth & $311.35 \pm 24.45$ & $28.75 \pm 1.68$ \\
 1 & $41.57 \pm 2.05$ & $4.18 \pm 0.03$ \\
 2 & $40.64 \pm 0.28$ & $4.07 \pm 0.09$ \\
 4 & $44.18 \pm 2.87$ & $4.44 \pm 0.21$ \\
 8 & $42.51 \pm 1.18$ & $4.28 \pm 0.13$ \\
 16 & $42.75 \pm 2.34$ & $4.22 \pm 0.17$\\
 32 (original) & $41.75 \pm 1.33$ & $4.14 \pm 0.02$ \\
 \bottomrule
\end{tabular}
\end{center}
\caption{Generative performance of GSN models trained with different depth resolutions. Remarkably, we show that we can down-sample depth resolution to \textit{a single pixel} without degrading the quality of the generative model.}
\label{tab:depth_resolution}
\end{table}

\textbf{Does $D_\phi$ need to be regularized?} Different forms of regularizing the discriminator affect the quality of generated samples. We discover that $D_\phi$ greatly benefits from regularization (Tab.~\ref{tab:regularizer_ablation}). In particular, data augmentation~\cite{diffaugment} and the discriminator reconstruction penalty~\cite{self_supervised_discriminator} are both crucial; training to rapidly diverge without either of these components. The R1 gradient penalty~\cite{r1} offers an additional improvement of training stability helping adversarial learning to converge.

\begin{table}[h]
\small
\begin{center}
 \begin{tabular}{lcc}
 \toprule
 Model & FID $\downarrow$ & SwAV-FID $\downarrow$ \\
 \midrule
 GSN (full model) & $41.75 \pm 1.33$ & $4.14 \pm 0.02$ \\
  \quad -- R1 gradient penalty & $52.9 \pm 19.9$ & $4.56 \pm 1.50$\\
 \quad -- Reconstruction penalty & $274.3 \pm 41.4$ & $23.58 \pm 4.11$\\
 \quad -- Data augmentation & $412.25 \pm 14.34$ &  $35.71 \pm 9.51$\\
 \bottomrule

\end{tabular}
\end{center}
\caption{Regularizing $D_\phi$ shows great benefits for GSN, especially when using a reconstruction penalty and data augmentation.}
\label{tab:regularizer_ablation}
\end{table}

\subsection{View Synthesis}

We now turn to the task of view synthesis, where we show how GSN performs in comparison with two approaches for scene-level view synthesis of free moving cameras: Generative Temporal models with Spatial Memory (GTM-SM) \cite{gtmsm} and Incremental Scene Synthesis (ISS) \cite{iss}. Taking the definition in \cite{iss}, the view synthesis task is defined as follows: for a given step $t$ in a sequence we want the model to predict the target $t+5$ views $\mathcal{T}=\{(\textbf{X},~\textbf{T})_i\}_{i=t:t+5}$ conditioned on the source $t-5$ views $\mathcal{S}=\{(\textbf{X},~\textbf{T})_i\}_{i=t-5:t}$ along the camera trajectory. Note that this view synthesis problem is unrelated to video prediction, since the scenes are static and camera poses for source and target views are given. To tackle this task both GTM-SM \cite{gtmsm} and ISS \cite{iss} rely on auto-encoder architectures augmented with memory mechanisms that directly adapt to the conditional nature of the task. For GSN to deal with this task we follow standard practices for GAN inversion \cite{ganinversion} (see supplementary material for details on the encoder architecture and inversion algorithm). We invert source views $\mathcal{S}$ into the prior and use the resulting latent to locally condition the radiance field and render observations using the camera poses of the target views $\mathcal{T}$.

Following \cite{iss} we report results on the Vizdoom \cite{vizdoom} and AVD \cite{avd} datasets in Tab.~\ref{tab:inversion_quant}.\footnote{There is no code or data release for \cite{iss}. In private communication with the authors of \cite{iss} we discussed the data splits and settings used for generating trajectories and follow them as closely as possible. We thank them for their help.} We report two different aspects of view synthesis: the capacity to fit the source views or \textit{Memorization} (\eg reconstruction), and the ability to predict the target views or \textit{Hallucination} (\eg scene completion) using L1 and SSIM metrics.\footnote{We note that while these metrics are a good proxy for reconstruction quality, they do not tell a complete picture in the hallucination setting due to the stochastic nature of the view synthesis problem.} GSN outperforms both GTM-SM \cite{gtmsm} and ISS \cite{iss} for nearly all tasks (Tab.~\ref{tab:inversion_quant}), even though it was not trained to explicitly learn a mapping from $\mathcal{S}$ to $\mathcal{T}$ (see supplementary material for qualitative results). We attribute this success to the powerful scene prior learned by GSN.

\begin{table}[!t]
\centering
\small
\subcaptionbox{View synthesis results on Vizdoom \cite{vizdoom}}{
 \begin{tabular}{lcccc} 
  \toprule
    &  \multicolumn{2}{c}{Memorization} & \multicolumn{2}{c} {Hallucination}\\
  \cmidrule(r){2-3}\cmidrule(r){4-5}
  & L1 $\downarrow$ & SSIM $\uparrow$ & L1 $\downarrow$ & SSIM $\uparrow$ \\ 
  \midrule
  GTM-SM\cite{gtmsm} & $0.09$ & $0.52$ & $0.13$ & $0.49$   \\
  ISS \cite{iss} & $0.09$ & $0.56$ & $\textbf{0.11}$ & $0.54$ \\
  GSN  & $\textbf{0.07}$ & $\textbf{0.66}$ & $\textbf{0.11}$ & $\textbf{0.57}$ \\
  \bottomrule
\end{tabular}
}
\vspace{0.3 cm}
\small
\subcaptionbox{View synthesis results on AVD\cite{avd}}{
 \begin{tabular}{lcccc} 
   \toprule
    &  \multicolumn{2}{c}{Memorization} & \multicolumn{2}{c} {Hallucination} \\
   \cmidrule(r){2-3}\cmidrule(r){4-5}
  & L1 $\downarrow$ & SSIM $\uparrow$ & L1 $\downarrow$ & SSIM $\uparrow$ \\ 
  \midrule
  GTM-SM\cite{gtmsm} & $0.37$ & $0.12$ & $0.43$ & $0.1$   \\
  ISS \cite{iss} & $0.22$ & $0.31$ & $\textbf{0.25}$ & $0.23$ \\
  GSN  & $\textbf{0.19}$ & $\textbf{0.54}$ & $0.35$ & $\textbf{0.35}$ \\
  \bottomrule
\end{tabular}
}
\caption{Results of view synthesis on Vizdoom \cite{vizdoom} (a) and AVD \cite{avd} (b). GSN improves view synthesis quality as a result of modeling a powerful prior over scenes.}
\label{tab:inversion_quant}
\end{table}

\section{Conclusions}

We have made great strides towards generative modeling of unconstrained, complex and realistic indoor scenes. We introduced GSN, a generative model that decomposes a scene into many local radiance fields, which can be rendered by a free moving camera. Our decomposition scheme scales efficiently to big scenes while preserving details and distribution coverage. We show that GSN can be trained on multiple scenes to learn a rich scene prior, while rendered trajectories on scenes sampled from the prior are smooth and consistent, maintaining scene coherence. The prior distribution learned by GSN can also be used to infer observations from arbitrary cameras given a sparse set of observations. A multitude of avenues for future work are now enabled by GSN, from improving the rendering performance and training on large-scale datasets to exploring the wide range of down-stream tasks that benefit from a learned scene prior like model-based reinforcement learning \cite{gamegan}, SLAM \cite{neuralslam1} or 3D completion for immersive photography \cite{nerfies}.

{\small
\bibliographystyle{ieee_fullname}
\bibliography{arxiv.bbl}
}

\clearpage
\appendix
\section{Model Architectures and Training Details} %
In this section we summarize the model architectures, hyperparameter settings, and other training details used for producing the GSN models presented in this paper.

\subsection{Mapping Network}
The mapping network maps the global latent code $\textbf{z}$ to an intermediate non-linear latent space~\cite{stylegan}. All models in our experiments, including those that do not use the global generator, use a mapping network which consists of a normalization step followed by three linear layers with LeakyReLU activations~\cite{leakyrelu}, as shown in Tab.~\ref{tab:mapping_network_architecture}.

\begin{table}[h]
	\small
	\centering
	\begin{tabular}{lcc}
		\toprule
		& Activation & Output Shape \\
		\midrule
		Input $\textbf{z}$ & -- & $128$ \\
		Normalize & -- & $128$ \\
		Linear & LeakyReLU (0.2) & $128$ \\
		Linear & LeakyReLU (0.2) & $128$ \\
		Linear & LeakyReLU (0.2) & $128$ \\
		\bottomrule
	\end{tabular}
	\caption{Mapping network architecture.}
	\label{tab:mapping_network_architecture}
\end{table}

\subsection{Global Generator}

The purpose of the global generator (Tab.~\ref{tab:global_generator_architecture}) is to map from a single global latent code $\textbf{z}$ to a 2D grid of local latent codes $\textbf{W}$ which represent the spatial layout of the scene. The global generator is composed of successive modulated convolutional layers~\cite{stylegan2}
that are conditioned on the global latent code.
Following StyleGAN~\cite{stylegan}, the model learns a constant input for the first layer. The first layers in every pair of modulated convolutional layers thereafter upsamples the feature map resolution by $2\times$, which is implemented as a transposed convolution with a stride of 2, followed by bilinear filtering~\cite{blur}. 

The output resolution of the global generator (which is also the spatial resolution of $\textbf{W}$) is a hyperparameter which effectively controls the size of the spatial region represented by each individual local latent code. We set the global generator output resolution to $32\times32$ for all experiments.

\begin{table}[h]
	\small
	\centering
	\begin{tabular}{lcc}
		\toprule
		& Activation & Output Shape \\
		\midrule
		Constant Input & -- & $256 \times 4 \times 4$ \\
		ModulatedConv ($3\times3$) & LeakyReLU (0.2) & $256 \times 8 \times 8$ \\
		ModulatedConv ($3\times3$) & LeakyReLU (0.2) & $256 \times 8 \times 8$ \\
		ModulatedConv ($3\times3$) & LeakyReLU (0.2) & $256 \times 16 \times 16$ \\
		ModulatedConv ($3\times3$) & LeakyReLU (0.2) & $256 \times 16 \times 16$ \\
		ModulatedConv ($3\times3$) & LeakyReLU (0.2) & $256 \times 32 \times 32$ \\
		ModulatedConv ($3\times3$) & LeakyReLU (0.2) & $256 \times 32 \times 32$ \\
		ModulatedConv ($3\times3$) & -- & $32 \times 32 \times 32$ \\
		\bottomrule
	\end{tabular}
	\caption{Global generator architecture.}
	\label{tab:global_generator_architecture}
\end{table}

\subsection{Local Generator}

The local generator is composed of a locally conditioned radiance field network which maps coordinates and view direction to appearance $a$ and occupancy $\sigma$, a volumetric rendering step which accumulates along sampled rays to convert $a$ and $\sigma$ values to feature vectors, and a refinement network which upsamples feature maps to higher resolution RGB images.

\begin{figure}[h]
	\centering
	\includegraphics[width=0.32\textwidth]{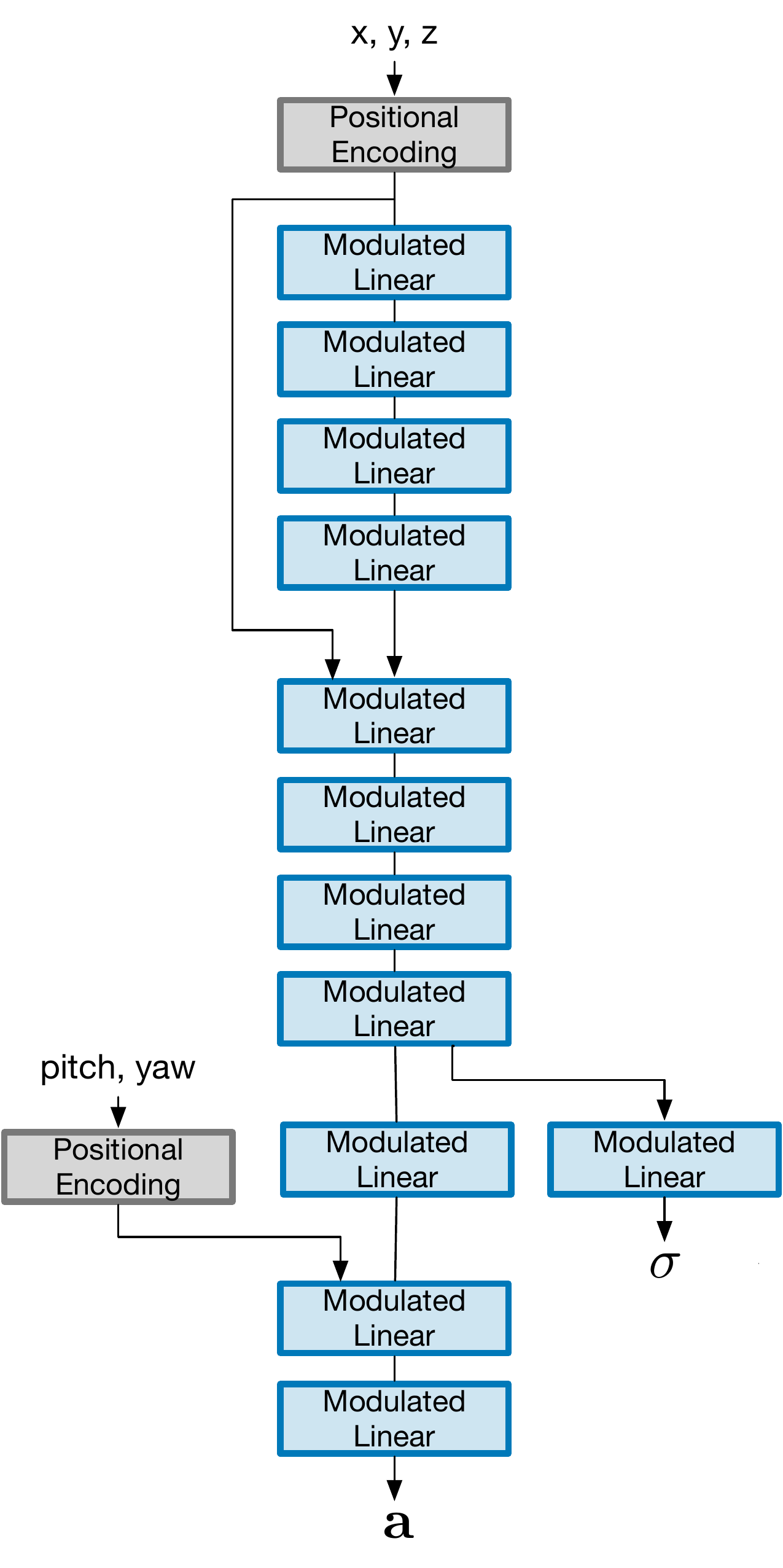}
	\caption{Locally conditioned radiance field network.}
	\label{fig:locally_conditioned_radiance_field_full}
\end{figure}

The locally conditioned radiance field network (Fig.~\ref{fig:locally_conditioned_radiance_field_full}) mimics the architecture of the the original NeRF network~\cite{nerf}. To condition the network such that it can represent many different radiance fields we swap out the fixed linear layers for modulated linear layers similar to those used in CIPS~\cite{modfc}, where each modulated linear layer is conditioned on $\mathbf{w_{ij}}$. Each modulated linear layer has 128 channels.

When performing volumetric rendering we threshold occupancy values $\sigma$ with a softplus as in D-NeRF~\cite{nerfies} as opposed to the standard ReLU, as we find it leads to more stable training. For all experiments we sample 64 samples per ray. When generating $64\times64$ images we sample the radiance field network to produce feature maps at $32\times32$ resolution, and when generating $128\times128$ resolution images we sample feature maps at $64\times64$ resolution.

Once volumetric rendering has been performed we upsample the resulting feature map with refinement blocks (Fig.~\ref{fig:residual_block}) until the desired resolution is achieved, then apply a sigmoid to bound the final output, as in GIRAFFE~\cite{giraffe}. In general, we found that sampling higher resolution feature maps directly from the radiance field produced higher quality results compared to sampling at low resolution and applying many refinement blocks, but the computational cost is significantly higher.

\begin{figure}[h]
	\centering
	\includegraphics[width=0.256\textwidth]{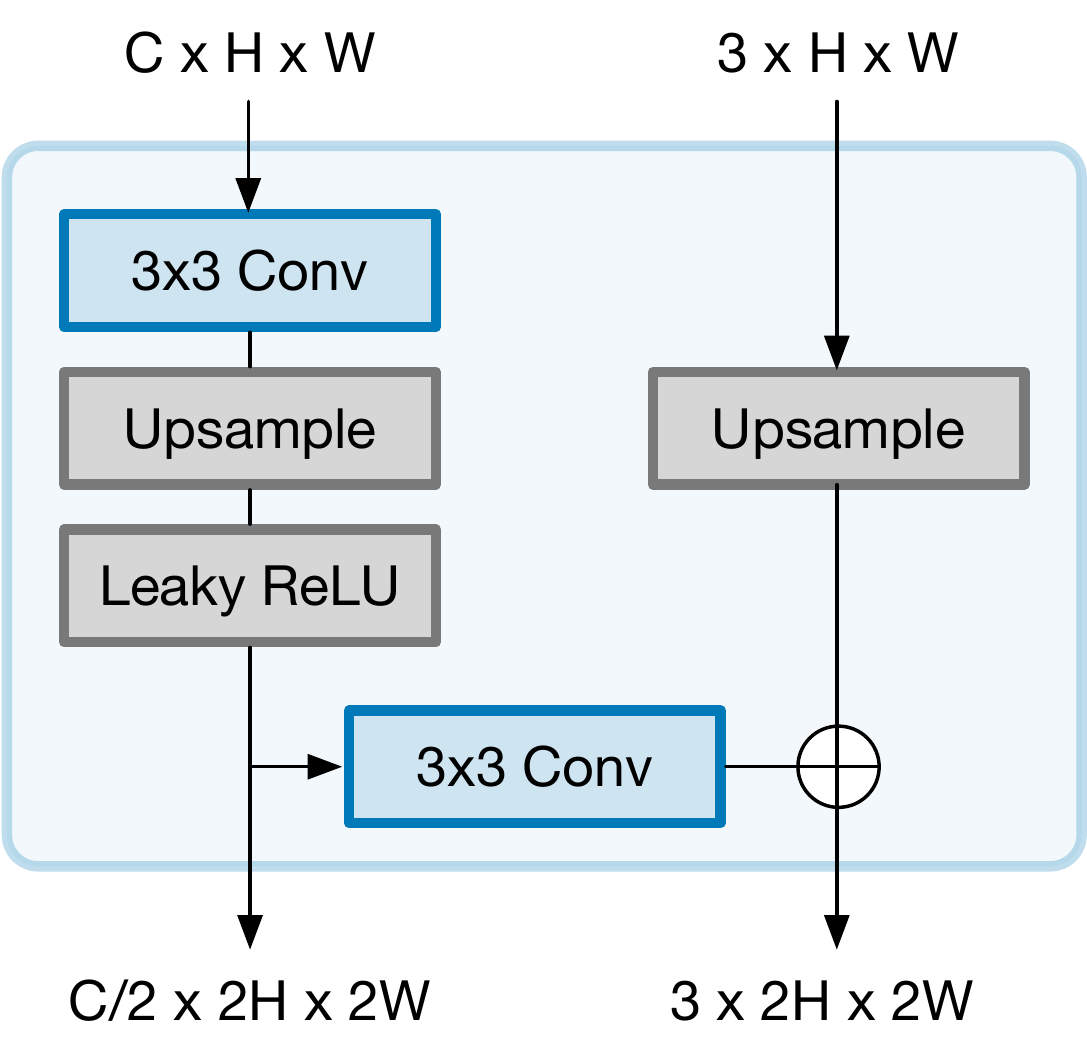}
	\caption{Refinement block. Feature maps pass through the left path, while RGB images pass through the right path. The input to the right path is not applied for the first refinement block after the volumetric rendering step.}
	\label{fig:refinement_block}
\end{figure}

\subsection{Discriminator}

\begin{figure}[h]
	\centering
	\includegraphics[width=0.22\textwidth]{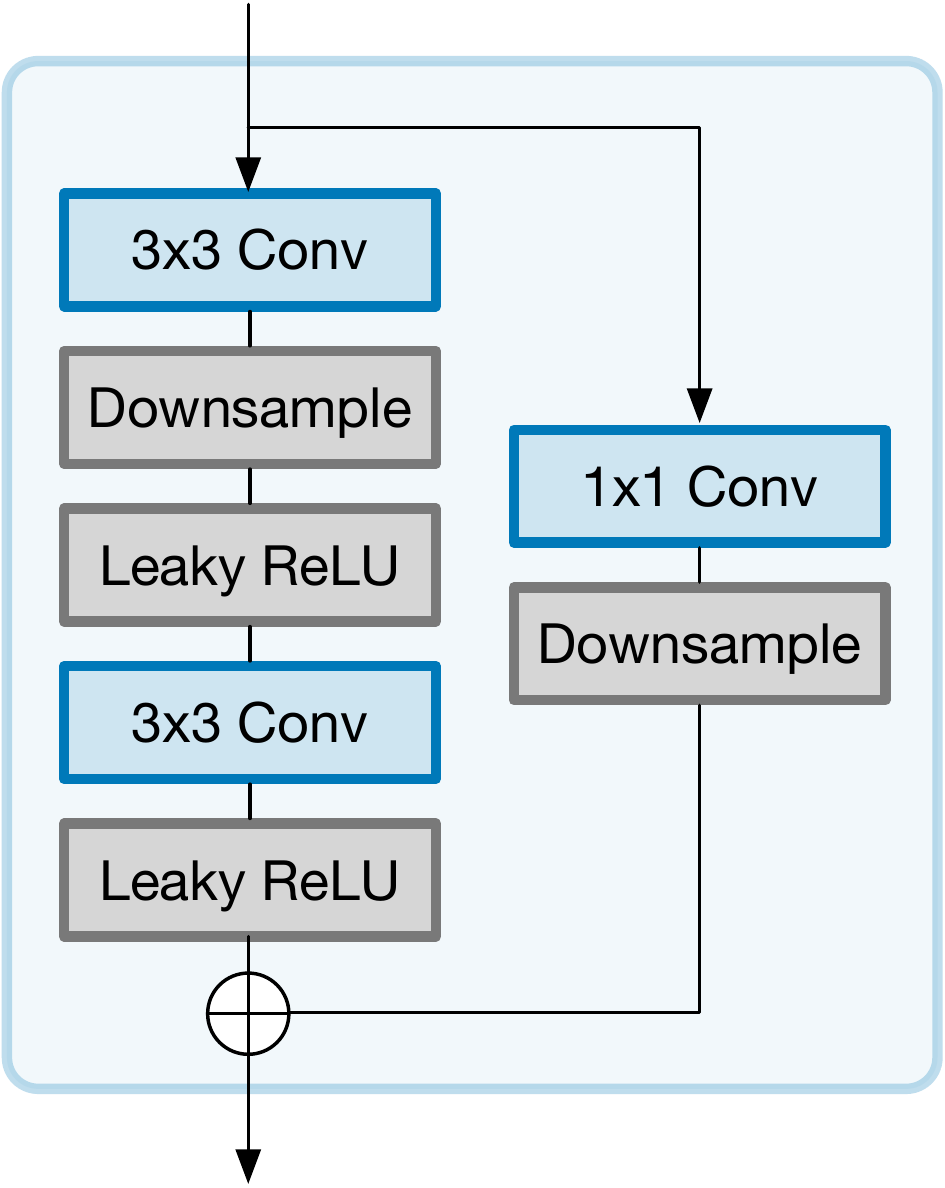}
	\caption{Residual block.}
	\label{fig:residual_block}
\end{figure}

The discriminator is based on the architecture used in StyleGAN2\cite{stylegan2} (Tab.~\ref{tab:discriminator_architecture}), including residual blocks (Fig.~\ref{fig:residual_block}) and minibatch standard deviation layer \cite{progan}. When including depth information as input to the discriminator we normalize it to $[0, 1]$. In the case that the radiance field network is sampled at a resolution lower than the final output resolution (such as when using the refinement network), then resulting depth maps will have lower resolution than real examples.
To prevent the discriminator from using this difference in detail to differentiate real and fake samples we downsample all real depth maps to match the resolution of the generated depth maps, then upsample them both back to full resolution. 

\begin{table}[h]
	\small
	\centering
	\begin{tabular}{lcc}
		\toprule
		& Activation & Output Shape \\
		\midrule
		Input RGB-D & -- & $4 \times 64 \times 64$ \\
		Conv ($3\times3$) & LeakyReLU (0.2) & $64 \times 64 \times 64$ \\
		Residual Block & LeakyReLU (0.2) & $128 \times 32 \times 32$ \\
		Residual Block  & LeakyReLU (0.2) & $256 \times 16 \times 16$ \\
		Residual Block & LeakyReLU (0.2) & $512 \times 8 \times 8$ \\
		Residual Block & LeakyReLU (0.2) & $512 \times 4 \times 4$ \\
		Minibatch stdev & -- & $513 \times 4 \times 4$ \\
		Conv ($3\times3$) & LeakyReLU (0.2) & $512 \times 4 \times 4$ \\
		Flatten & --& $8192$ \\
		Linear & LeakyReLU (0.2) & $512$ \\
		Linear & -- & $1$\\
		\bottomrule
	\end{tabular}
	\caption{Discriminator architecture.}
	\label{tab:discriminator_architecture}
\end{table}

The decoder (Tab.~\ref{tab:decoder_architecture}) takes as input $4\times4$ resolution feature maps from the discriminator (before the minibatch standard deviation layer), and applies successive transposed convolutions with a stride of 2 and bilinear filtering~\cite{blur} to upsample the input until the original resolution is recovered.

\begin{table}[h]
	\small
	\centering
	\begin{tabular}{lcc}
		\toprule
		& Activation & Output Shape \\
		\midrule
		Input Feature Map & -- & $512 \times 4 \times 4$ \\
		ConvTranspose ($3\times3$) & LeakyReLU (0.2) & $256\times 8 \times 8$ \\
		ConvTranspose ($3\times3$) & LeakyReLU (0.2) & $128\times 16 \times 16$ \\
		ConvTranspose ($3\times3$) & LeakyReLU (0.2) & $64\times 32 \times 32$ \\
		ConvTranspose ($3\times3$) & LeakyReLU (0.2) & $32\times 64 \times 64$ \\
		Conv ($3\times3$) & -- & $4\times 64 \times 64$ \\
		\bottomrule
	\end{tabular}
	\caption{Decoder architecture}
	\label{tab:decoder_architecture}
\end{table}

\subsection{Sampling Camera Poses}
We use poses from  camera trajectories in the training set as candidate poses during sampling, as real camera poses better reflect the true distribution of occupable locations compared to uniformly sampling over the entire scene region. Sampled camera poses are normalized and expressed relative to the camera pose in the middle of the trajectory. This normalization enforces an egocentric coordinate system whose origin is placed at the center of $\textbf{W}$. Note that despite working with trajectories of multiple camera poses, we still only sample a single camera pose per generated scene during training.

\subsection{Training Details} %
We use the RMSprop~\cite{rmsprop} optimizer with $\alpha=0.99$, $\epsilon=10^{-8}$, and a learning rate of 0.002 for both the generator and discriminator. Following StyleGAN~\cite{stylegan}, we set the learning rate of the mapping network $100\times$ less than the rest of the network for improved training stability. Equalized learning rate~\cite{progan} is used for all learnable parameters, and an exponential moving average of the generator weights~\cite{progan} with a decay of 0.999 is used to stabilize test-time performance. Differentiable data augmentations~\cite{diffaugment} such as random translation, colour jitter, and Cutout~\cite{cutout} are applied to all inputs to the discriminator in order to combat overfitting. To save compute, the R1 gradient penalty is applied using a lazy regularization strategy~\cite{stylegan2} by applying it every 16 iterations. We set $\lambda_{R1}$ to 0.01 and $\lambda_{Recon}$ to 1000 for all experiments.

All $64\times64$ resolution models used for the generation performance evaluation (GSN and otherwise) were trained for 500k iterations with a batch size of 32. Training takes 4 days on two NVIDIA A100 GPUs with 40GB of memory each. Mixed precision training is applied to the generator for a small reduction in memory cost and training time. We do not apply mixed precision training to the discriminator as training stability decreases in this case.

\section{Inverting GSN for View Synthesis}

In order for GSN to deal with the view synthesis problem we follow common practices for GAN inversion \cite{ganinversion}, adopting a hybrid inversion approach where we first train an encoder $E_{\theta_E}: \mathbb{R}^{3 \times w \times h}\times SE(3) \longrightarrow \mathbb{R}^{c \times s \times s \times s}$ on $\{(\hat{\textbf{X}}, \textbf{T}, \textbf{W})_i\}_{i=1:n}$ tuples sampled from a trained GSN (trained on the training set of the same dataset). The goal of this encoder is to predict an initial grid of latent codes $\textbf{W}_0$ given a set of posed views. Our encoder is conceptually similar to \cite{mvsmachine,atlas} where views $\hat{\textbf{X}}$ are first processed with a backbone (UNet \cite{unet} with a ResNet-50 encoder in our case) and the resulting feature maps are back-projected using camera poses $\textbf{T}$ into a shared feature volume $\textbf{V} \in \mathbb{R}^{c \times s \times s \times s}$. Finally, we perform average pooling over the height dimension of $\textbf{V}$ to get $\textbf{W}_0 \in \mathbb{R}^{c \times s \times s}$. We train the encoder by minimizing the following reconstruction loss:

\begin{eqnarray}
	\mathcal{L}(\hat{\textbf{X}}, \textbf{T}, \textbf{W}; \theta_E) =\|\textbf{W} - E_{\theta_E}(\hat{\textbf{X}}, \textbf{T}) \|_2 + \\
	+ \|\hat{\textbf{X}} - f(E_{\theta_E}(\hat{\textbf{X}}, \textbf{T}), \textbf{T}) \|_2,
\end{eqnarray}

where the first term encourages the reconstruction of the local latent grid, and the second term encourages samples from locally conditioned radiance field $f$ to match the original input views.

\begin{figure*}[h]
	\centering
	\includegraphics[width=\linewidth]{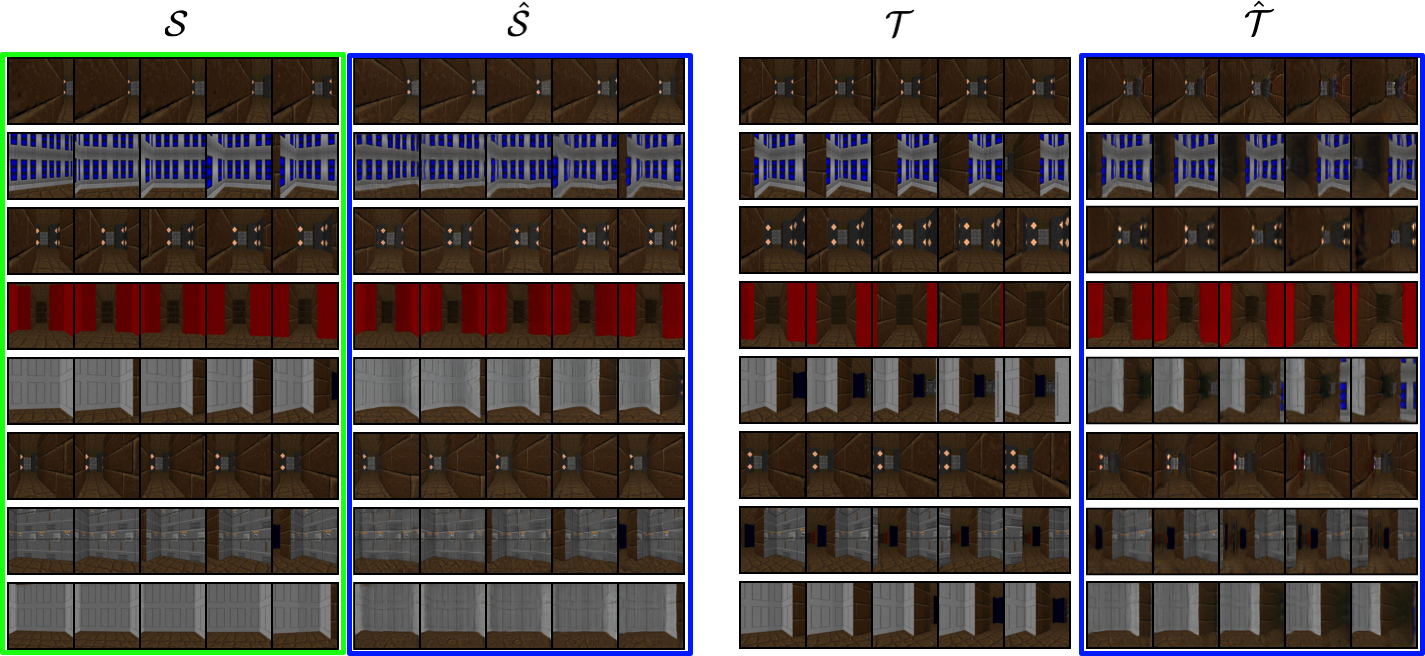}
	\caption{Qualitative view synthesis results on Vizdoom sequences not seen during training. Given source views $\mathcal{S}$ we invert GSN to obtain a local latent code grid $\hat{\textbf{W}}$, which is then use both to reconstruct $\mathcal{S}$, denoted as $\hat{\mathcal{S}}$, and also to predict target views $\mathcal{T}$ (given their camera poses) which are denoted as $\hat{\mathcal{T}}$. Each row corresponds to a different set of source views $\mathcal{S}$. Frames highlighted in green are input to GSN, frames highlighted in blue are predictions.}
	\label{fig:qual_view_syn}
\end{figure*}

At inference time, given a trained encoder $E_{\theta_E}$, we feed the source views $\mathcal{S}=\{(\textbf{X}, \textbf{T})_i\}_{i=t-5:t}$ through our encoder to predict an initialization latent code grid $\textbf{W}_0$ that we then optimize via SGD for $1000$ iterations to get $\hat{\textbf{W}}$. Given that scenes do not share a canonical orientation, we predict $\textbf{W}_0$ at multiple rotation angles of $\{\textbf{T}_i\}_{i=t-5:t}$ about the $y-$axis to find the generator's preferred orientation and use this orientation during optimization (note that relative transformations between camera poses do not change with this global transformation). We define the preferred orientation as the one that minimizes an auto-encoding LPIPS loss \cite{lpips}. The optimization process is performed by freezing the weights of $f_{\theta_f}$ and computing a reconstruction loss \wrt $\mathcal{S}$. We then use $\hat{\textbf{W}}$ in the locally conditioned radiance field and render observations using the camera poses of $\mathcal{S}$ to produce $\hat{\mathcal{S}}$ (\ie to auto-encode source views), while also rendering from the camera poses of the target views $\mathcal{T}$ to produce $\hat{\mathcal{T}}$ (\ie to predict unseen parts of the scene). Future work will explore in depth how to improve the quality and efficiency of the inversion approach for GSN-based models where the generative model tends to prefer a certain orientation. In Fig. \ref{fig:qual_view_syn} we show qualitative results for view synthesis on Vizdoom on held out sequences not seen during training. We can see how GSN learns a robust prior that is able to fill in the scene with plausible completions (\eg row 5 and 8), even if those completions do not strictly minimize the L1 reconstruction loss. 

\begin{figure*}[h]
	\centering
	\includegraphics[width=\linewidth]{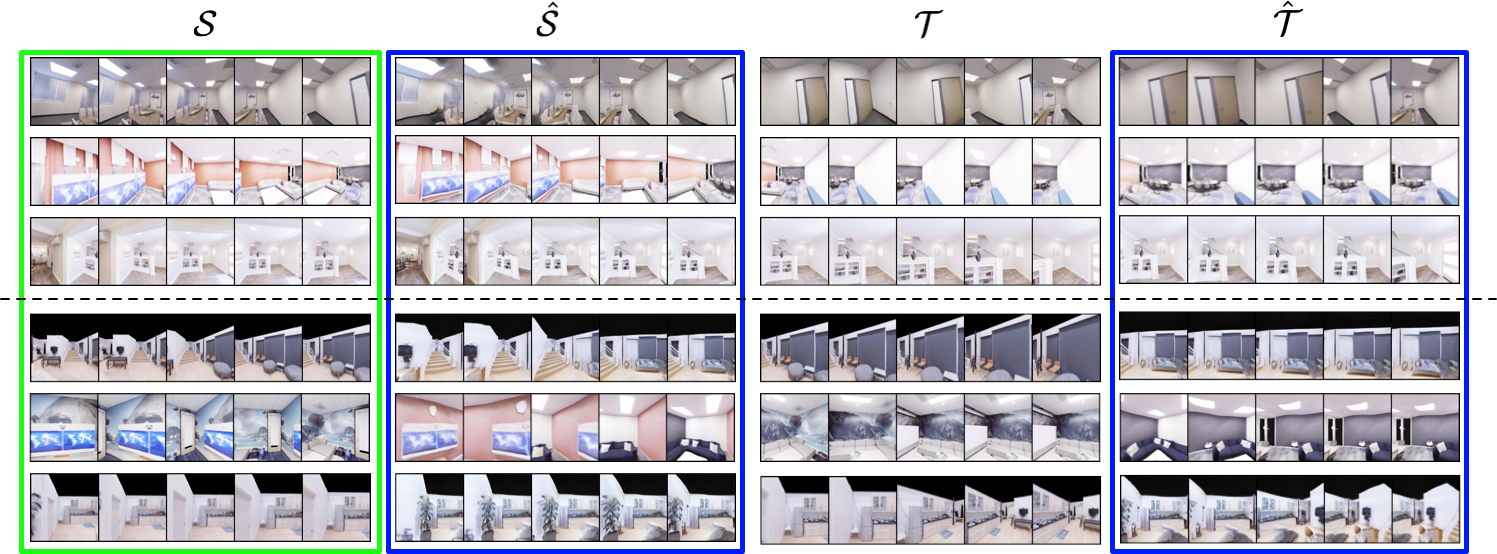}
	\caption{Qualitative view synthesis results on Replica. Given source views $\mathcal{S}$ we invert GSN to obtain a local latent code grid $\hat{\textbf{W}}$, which is then use both to reconstruct $\mathcal{S}$, denoted as $\hat{\mathcal{S}}$, and also to predict target views $\mathcal{T}$ (given their camera poses) which are denoted as $\hat{\mathcal{T}}$. Each row corresponds to a different set of source views $\mathcal{S}$ (top 3 rows are scenes from the training set, bottom 3 rows are scenes in a heldout test set). Frames highlighted in green are input to GSN, frames highlighted in blue are predictions.}
	\label{fig:qual_view_syn_replica}
\end{figure*}

In addition, Fig. \ref{fig:qual_view_syn_replica} shows qualitative view synthesis results on the Replica dataset \cite{replica}, showing the applicability of GSN for view synthesis on realistic data. In this experiment we follow the settings described for Vizdoom in terms of $\mathcal{S}$ and $\mathcal{T}$. In Fig. \ref{fig:qual_view_syn_replica} the top 3 rows show results on data from the training set (\eg scenes that were observed during training) and the bottom 3 rows show test set results (\eg results on unseen scenes). We can see how GSN successfully uses the prior learned from training data to find a plausible scene completion for unseen scenes that respects the global scene geometry.

\section{Qualitative Results on Local vs. Global Coordinate Systems} %

\begin{figure*}[h]
	\centering
	\begin{subfigure}{0.49\linewidth}
		\centering
		\includegraphics[width=\linewidth]{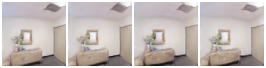}
		\includegraphics[width=\linewidth]{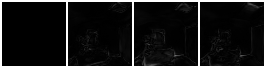}
		\includegraphics[width=\linewidth]{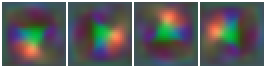}
	\end{subfigure} \hfill
	\begin{subfigure}{0.49\linewidth}
		\centering
		\includegraphics[width=\linewidth]{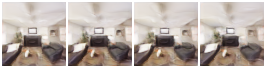}
		\includegraphics[width=\linewidth]{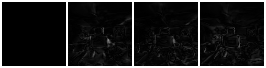}
		\includegraphics[width=\linewidth]{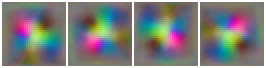}
	\end{subfigure}
	\caption{Change in generation output as local latent codes are rotated with a \textit{local} coordinate system for two different scenes. (Top) Rendered image. (Middle) Residual \wrt 0 degree rotation. (Bottom) Visualization of $\textbf{W}$. Each column corresponds to a rotation of the camera and $\textbf{W}$ in $\{0, 90, 180, 270\}$.}
	\label{fig:rotate_W_local_fix_camera}
\end{figure*}

\begin{figure*}[h]
	\centering
	\begin{subfigure}{0.49\linewidth}
		\centering
		\includegraphics[width=\linewidth]{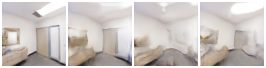}
		\includegraphics[width=\linewidth]{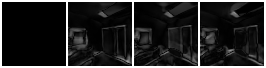}
		\includegraphics[width=\linewidth]{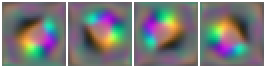}
		
	\end{subfigure} \hfill
	\begin{subfigure}{0.49\linewidth}
		\centering
		\includegraphics[width=\linewidth]{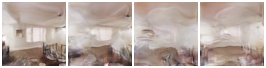}
		\includegraphics[width=\linewidth]{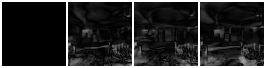}
		\includegraphics[width=\linewidth]{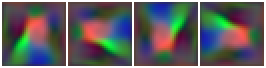}
	\end{subfigure}
	\caption{Change in generation output as local latent codes are rotated with a \textit{global} coordinate system. (Top) Rendered image. (Middle) Residual \wrt 0 degree rotation. (Bottom) Visualization of $\textbf{W}$. Each column corresponds to a rotation of the camera and $\textbf{W}$ in $\{0, 90, 180, 270\}$.}
	\label{fig:rotate_W_global_fix_camera}
\end{figure*}

\begin{figure*}[h]
	\centering
	\begin{subfigure}{0.49\linewidth}
		\centering
		\includegraphics[width=\linewidth]{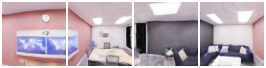}
		\includegraphics[width=\linewidth]{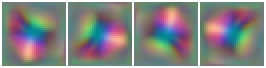}
	\end{subfigure} \hfill
	\begin{subfigure}{0.49\linewidth}
		\centering
		\includegraphics[width=\linewidth]{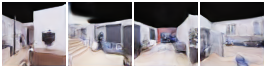}
		\includegraphics[width=\linewidth]{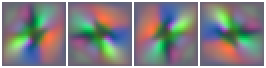}
	\end{subfigure}
	\caption{Change in generation output for a \textit{fixed camera} as local latent codes are rotated with a \textit{local} coordinate system for two different scenes. (Top) Rendered image. (Bottom) Visualization of $\textbf{W}$. Each column corresponds to a rotation of $\textbf{W}$ in $\{0, 90, 180, 270\}$.}
	\label{fig:rotate_W_local}
\end{figure*}

\begin{figure*}[h]
	\centering
	\begin{subfigure}{0.49\linewidth}
		\centering
		\includegraphics[width=\linewidth]{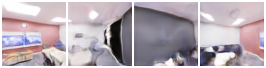}
		\includegraphics[width=\linewidth]{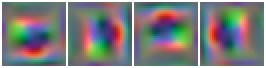}
	\end{subfigure} \hfill
	\begin{subfigure}{0.49\linewidth}
		\centering
		\includegraphics[width=\linewidth]{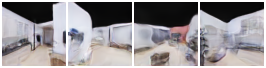}
		\includegraphics[width=\linewidth]{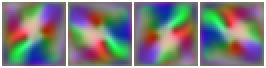}
	\end{subfigure}
	\caption{Change in generation output for a \textit{fixed camera} as local latent codes are rotated with a \textit{global} coordinate system for two different scenes. (Top) Rendered image. (Bottom) Visualization of $\textbf{W}$. Each column corresponds to a rotation of $\textbf{W}$ in $\{0, 90, 180, 270\}$.}
	\label{fig:rotate_W_global}
\end{figure*}

In this section we demonstrate the robustness of GSN \wrt re-arrangement of the latent codes in $\textbf{W}$. In order to do so we sample different scenes from our learned prior and apply a rigid transformation to their corresponding $\textbf{W}$ (a 2D rotation). In principle, this rotation should amount to a rotation of the scene represented by $\textbf{W}$ that does not change the radiance field prediction. To qualitatively evaluate this effect we sample different scenes and rotate their corresponding $\textbf{W}$ by $\{0, 90, 180, 270\}$ degrees while (i) rotating the camera by the same amount about the $y-$axis so that the rendered image should remain constant and (ii) keeping the camera fixed so that the scene should rotate. In Fig. \ref{fig:rotate_W_local_fix_camera}-\ref{fig:rotate_W_global_fix_camera} we show the result of the setting in (i) for a local and global coordinate system respectively. In these results we see how a local coordinate system is drastically more robust to re-arrangements of the latent codes than a global coordinate system. In addition, we show results for the setting in (ii) in Fig. \ref{fig:rotate_W_local}-\ref{fig:rotate_W_global} for local and global coordinate systems respectively. In this case, we see how given a fixed camera, a rotation of $\textbf{W}$ amounts to rotating the scene. In this case we can also see how a local coordinate system results in higher rendering quality compared to that of a global coordinate system, which suffers from degradation as the rotation angle increases.

\section{Scene Editing} %

A nice property of the local latent grid produced by GSN is that it can be used to perform scene editing by directly altering $\textbf{W}$. This property allows us a degree of manual control for scene synthesis beyond what we get from randomly sampling the generator. While the low resolution of $\textbf{W}$ used in current models currently limits us to high level scene modifications, training with larger local latent grids could allow for more fine-grained control over scenes, such as rearrangement of furniture. 

\begin{figure}[h]
	\centering
	\setlength\tabcolsep{0.2em}
	\begin{tabular}{cc}
		\includegraphics[height=0.2\linewidth]{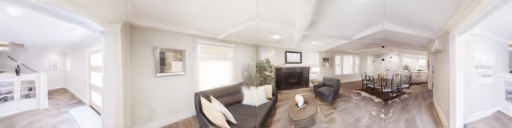} &
		\adjincludegraphics[height=0.2\linewidth, trim={0 0 {0.67\width} 0},clip]{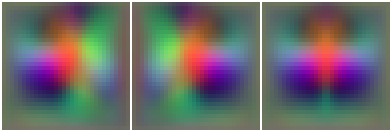}
	\end{tabular}

	\setlength\tabcolsep{0.2em}
	\begin{tabular}{cc}
		\includegraphics[height=0.2\linewidth]{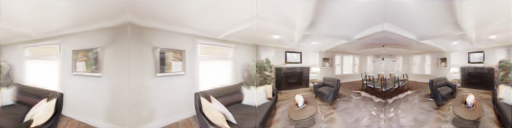} &
		\adjincludegraphics[height=0.20\linewidth, trim={{0.67\width} 0 0  0},clip]{figs/scene_editing/living_room/local_latents.png}
	\end{tabular}
	
	\caption{Panoramas and corresponding local latent codes for scenes produced by GSN. Mirroring the local latent code from a single room (top row) produces a new room (bottom row).}
	\label{fig:scene_edit_living_room}
\end{figure}

\begin{figure}[h]
	\centering
	\setlength\tabcolsep{0.2em}
	\begin{tabular}{cc}
		\includegraphics[height=0.2\linewidth]{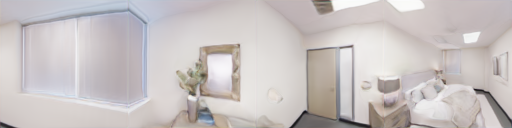} &
		\adjincludegraphics[height=0.2\linewidth, trim={0 0 {0.67\width} 0},clip]{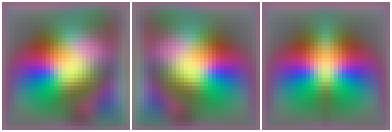}
	\end{tabular}

	\setlength\tabcolsep{0.2em}
	\begin{tabular}{cc}
		\includegraphics[height=0.2\linewidth]{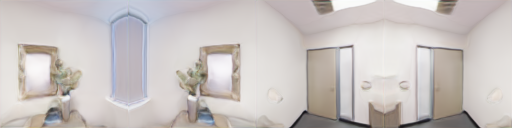} &
		\adjincludegraphics[height=0.20\linewidth, trim={{0.67\width} 0 0  0},clip]{figs/scene_editing/bedroom/local_latents.png}
	\end{tabular}
	
	\caption{Panoramas and corresponding local latent codes for scenes produced by GSN. Mirroring the local latent code from a single room (top row) produces a new room (bottom row).}
	\label{fig:scene_edit_bedroom}
\end{figure}

We find that, as with most image composition operations, the results of scene editing appear most convincing when the inputs are well aligned in terms of appearance and geometry. We demonstrate editing operations by manipulating the codes from single scenes, since we don't need to worry about matching appearance and geometry, but multiple scenes could be combined if they were similar enough.
In Fig.~\ref{fig:scene_edit_living_room}-\ref{fig:scene_edit_bedroom} we manipulate the local latent codes by mirroring them along the horizontal axis to produce unique scenes.

\end{document}